\documentclass[conference]{IEEEtran}
\IEEEoverridecommandlockouts
\usepackage{cite}
\usepackage{multirow}
\usepackage{amsmath,amssymb,amsfonts}
\usepackage{algorithmic}
\usepackage{tabularx}
\usepackage{graphicx}
\usepackage{adjustbox}
\usepackage{textcomp}
\usepackage{xcolor}
\usepackage{subcaption}
\usepackage{booktabs}
\usepackage{tikz}
\usepackage{changes}
\usepackage{makecell}
\usepackage{threeparttable}
\usepackage{hyperref}
\def\BibTeX{{\rm B\kern-.05em{\sc i\kern-.025em b}\kern-.08em
    T\kern-.1667em\lower.7ex\hbox{E}\kern-.125emX}}
\begin{document}

\title{SUPER: Seated Upper Body Pose Estimation using mmWave Radars\\
}

\newcommand{\linebreakand}{%
  \end{@IEEEauthorhalign}
  \hfill\mbox{}\par
  \mbox{}\hfill\begin{@IEEEauthorhalign}
}

%\author{\IEEEauthorblockN{
%Bo Zhang}
%\IEEEauthorblockA{\textit{Computing and Software Department} \\
%\textit{McMaster University)}\\
%Hamilton, Canada \\
%zhanb59@mcmaster.ca}
%\and
%\IEEEauthorblockN{
%Zimeng Zhou}
%\IEEEauthorblockA{\textit{Computing and Software %Department} \\
%\textit{McMaster University}\\
%Hamilton, Canada \\
%zhouz287@mcmaster.ca}
%\and
%\IEEEauthorblockN{ 
%Boyu Jiang}
%\IEEEauthorblockA{\textit{Computing and Software Department} \\
%\textit{McMaster University}\\
%Hamilton, Canada \\
%jiangb11@mcmaster.ca}
%\and

%\linebreakand
\author{
\IEEEauthorblockN{Bo Zhang, Zimeng Zhou, Boyu Jiang, Rong Zheng}
\IEEEauthorblockA{\textit{Department of Computing and Software} \\
\textit{McMaster University}\\
Hamilton, ON, Canada \\
{ \{{\it zhanb59, zhouz287, jiangb11, rzheng}\}}@mcmaster.ca}
}

\maketitle

\begin{abstract}
In industrial countries, adults spend a considerable amount of time sedentary each day at work, driving and during activities of daily living. Characterizing the seated upper body human poses using mmWave radars is an important, yet under-studied topic with many applications in human-machine interaction, transportation and road safety. In this work, we devise SUPER, a framework for seated upper body human pose estimation that utilizes dual-mmWave radars in close proximity. A novel masking algorithm is proposed to coherently fuse data from the
radars to generate intensity and Doppler point clouds with complementary information for high-motion but small radar cross section areas (e.g., upper extremities) and low-motion but large RCS areas (e.g. torso). A lightweight neural network extracts both global and local features of upper body and output pose parameters for the Skinned Multi-Person Linear (SMPL) model. Extensive leave-one-subject-out experiments on various motion sequences from multiple subjects show that SUPER outperforms a state-of-the-art baseline method by 30 -- 184\%. We also demonstrate its utility in a simple downstream task for hand-object interaction.
\end{abstract}

\begin{IEEEkeywords}
Seated upper body pose estimation, mmWave radars, data fusion, point clouds, deep neural networks
\end{IEEEkeywords}

\section{Introduction}
\label{sect:intro}
% why is upper body pose estimation interesting?
% interaction with everyday objects
% Gesture/sign languages
% driver distraction and fatigue
% what has been done in literature? (briefly)
Human pose estimation (HPE) estimates the configuration of human body parts from input data captured by sensors and has attracted much attention in industry and the research community due to its wide range of applications, including the human-machine interactions~\cite{liu2022egox}, fitness~\cite{wang2019aicoach}, virtual reality~\cite{anvari2022vrsurvey}, smart home~\cite{zhou2023metafi++} and smart vehicle~\cite{shinko2006tianalysis}, etc. 
% more detailed subdomains, diverse sensing systems, and advanced technologies in this field has been developed in recent years. 
While full-body HPE is important in characterizing joint movements during locomotions, a 2019 study showed that adults ages 20 to 75 in the US reported spending an average of 9.5 hours sedentary each day~\cite{matthews2021sedentary}. Therefore, seated upper body human pose estimation (SUB-HPE) is arguably more relevant in interactive applications and understanding users' mental states (e.g., alertness and attention). For example, by monitoring upper limb movements while sitting, novel applications can be developed to empower users to control digital interfaces, manipulate augmented reality environments, and manage smart home systems. SUB-HPE can also find applications in transportation and road safety, where drowsy or inattentive drivers pose a significant risk on roadways. Analyzing head poses, hand placements and orientation of the upper body allows the detection of early signs of drowsiness or distraction. 

% System-wise, many systems and processing pipelines are build around different sensors, including RGB cameras, depth cameras, Lidar sensors, acoustic sensors, and radar sensors.  

In recent years, the rapid advancements in deep learning led to significant progress in human body modeling~\cite{loper2015SMPL,osman2020star} and HPE using various sensing modalities. Notable work in HPE includes OpenPose~\cite{cao2019openpose} and VitPose~\cite{xu2022vitpose} in computer vision, Deep inertial poser~\cite{huang2018deep} and IMUPoser~\cite{mollyn2023imuposer} using  IMU sensors, mmPose~\cite{sengupta2020mm} and mmMesh~\cite{xue2021mmmesh} with mmWave radars, and DensePose~\cite{geng2022densepose} using WiFi devices, to name a few. 
Among different sensing modalities, mmWave radars offer distinct advantages due to their ability to penetrate obstructions like garments or walls, adapt to diverse lighting and weather conditions, and preserve user privacy. Furthermore, the substantial bandwidth (in the GHz range) equips mmWave radars with resilience against noise, interference, and center-meter level range resolutions. However, existing mmWave-based solutions predominantly target full-body locomotions and are not designed for handling nuanced upper-body movements. mmWave-based SUB-HPE shares with full-body HPE common challenges stemming from low spatial resolutions as the result of few on-board transmitting and receiving antennas on low-end commercial-of-the-shelf (COTS) mmWave radars, specular reflections and variations from inherent micro-body movements. But, crucially, it must also handle limited mobility in the upper body's core area when sitting, as well as the small radar cross-sections (RCS) of upper extremities, ranging from -45 dBsm to -20 dBsm for hands~\cite{hugler2016rcs}.

In this work, we devise SUPER, a framework for \underline{S}eated \underline{U}pper Body \underline{P}ose \underline{E}stimation using mmWave \underline{R}adars. The framework encompasses a dual-radar pre-processing and fusion pipeline and a light weight neural network to predict upper body pose parameters. 
To increase the spatial resolution of the acquired radar data, two closely positioned radar sensors, oriented perpendicular to each other, are utilized. A novel dual-radar masking algorithm coherently fuses data from the radars to generate two complementary types of point clouds:  the intensity point cloud (IPC) and the Doppler point cloud (DPC). The latter captures motion information of extremities while the former better characterizes low-motion portions of the upper body (e.g., torso areas). 
Benefiting from the sparse point cloud representation, the lightweight neural network extracts both global and local features of the upper body. Finally, the Skinned Multi-Person Linear (SMPL) model is applied to yield realistic human body poses and motions. An example of the data captured by an RGB camera, a motion capture system, and the predicted and ground truth poses can be found in Figure~\ref{fig:exampleofsuper}. 

\begin{figure*}[th]
    \centering
    \begin{subfigure}[b]{0.32\textwidth}
        \centering
        \includegraphics[height=1.4in]{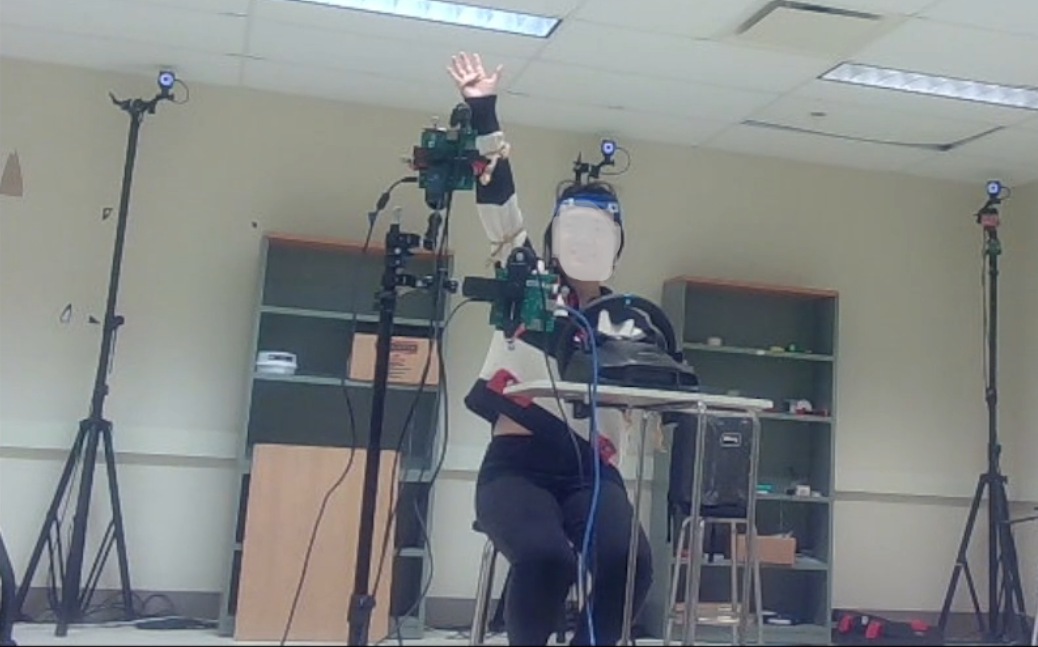}
        \caption{Camera view.}
        \label{fig:camera}
    \end{subfigure}
    %\hfill
    \begin{subfigure}[b]{0.32\textwidth}
        \centering
        \includegraphics[height=1.4in]{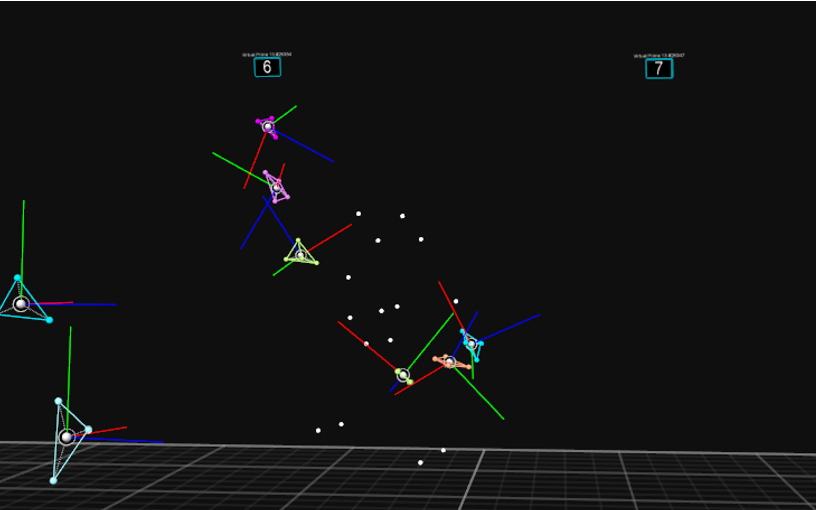}
        \caption{OptiTrack motion capture view.}
        \label{fig:optitrack}
    \end{subfigure}
    %\hfill
    \begin{subfigure}[b]{0.32\textwidth}
        \centering
        \includegraphics[height=1.4in]{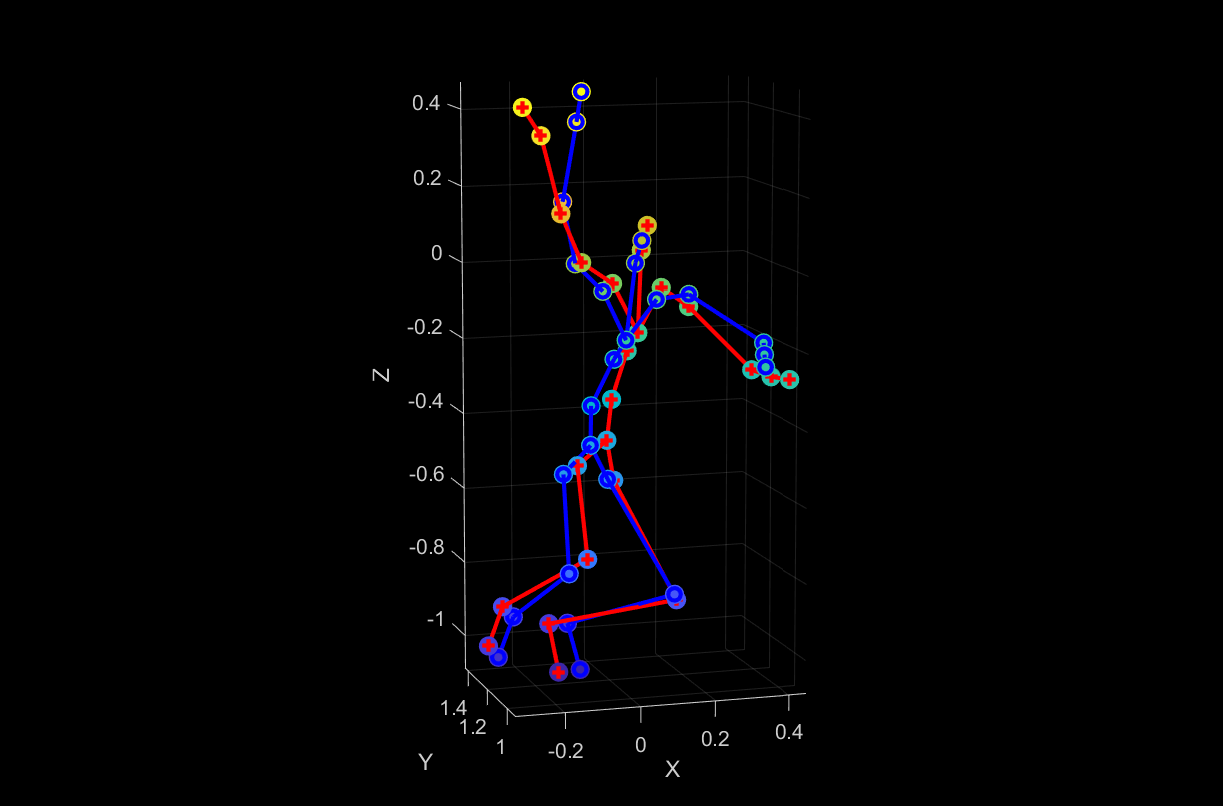}
        \caption{SUPER output}
        \label{fig:super}
    \end{subfigure}
\caption{The estimated skeleton model from SUPER vs. ground truth when a subject raises her/his hand up while seating. The blue circle markers stand for the estimated skeleton model, and the red plus markers are the corresponding ground truth.}
\label{fig:exampleofsuper}
\end{figure*}

We have implemented a prototype of SUPER utilizing two Texas Instruments IWR6843ISK mmWave radars\footnote{Demonstration videos can be found at \url{https://super-2023-web.github.io/SUPER/}.}. A diverse group of 10 subjects, encompassing different genders, ages, and body mass indices (BMIs), were recruited for data collection in a laboratory setting. The data collection process involved subjects engaging in predefined arm, head, torso motion sequences. Experiment results show that SUPER consistently outperforms a state-of-the-art (SOTA) baseline method and achieves 112mm in average Mean Per Joint Position Error (MPJPE) and 15.89mm Procrustes alignment MPJPE (PA-MPJPE) metrics in leave-one-subject out trials. To demonstrate the utility of SUPER, we also implement and evaluate a simple downstream task of hand-object interaction.

In summary, we make the following new contributions toward mmWave-based fine-grained SUB-HPE in this work.  
% contributions
\begin{itemize}
\item In this work, we investigate a new task, i.e., SUB-HPE, and collect a dataset consisting of various head, torso as well as arm motions using mmWave radars. 
\item The proposed framework, SUPER,  utilizes the intensity information from multi-antenna radar systems, to characterize the spatial occupation of human body under low mobility and Doppler information to capture motions of extremities. 
\item We demonstrate the feasibility of deploying two asynchronous but closely located mmWave radars to improve spatial resolution. A novel masking algorithm is proposed to coherently fuse data from both radars.
\item SUPER has been evaluated using different motion sequences and data from a diverse set of users and shows superior performance compared to a SOTA baseline method. 
\end{itemize}

% Organization of the paper
The rest of the paper is organized as follows. A review of recent development of mmWave-based HPE methods and public datasets is presented in Section \ref{sect:related}. In Section \ref{sect:method}, we introduce the proposed pipeline and key techniques. Section \ref{sect:impl} provides experiment setups and the dataset we build. Detailed results and system performance are provided in Section \ref{sect:eval}. Section \ref{sect:app} demonstrates the potentials of the proposed system by a downstream task.
Finally, we discuss the limitations of the work and conclude this paper in Section \ref{sect:conclusion}.

\section{Related work}
\label{sect:related}
FMCW radars as an emerging technology have attracted significant attention and have been investigated in a variety of sensing tasks, e.g. tracking and localization\cite{gu2019mmsense,wu2020mmtrack,zhao2019mid}, gesture recognition\cite{lien2016soli,liu2021m,palipana2021pantomime,khamis2020rfwash}, and vital sign monitoring\cite{yang2016monitoring,zhao2020heart,wang2021mmhrv,zhang2023pi}, etc. In this section, we focus on mmWave-based HPE methods and public datasets. 
\subsection{MmWave-based human pose estimation}
In \cite{sengupta2020mm}, Sengupta {\it et al.} present mm-Pose, which is among the first works in mmWave-based full-body HPE. mm-Pose projects radar point clouds from two separate and perpendicularly oriented radars onto the depth-azimuth(XY) and depth-elevation(XZ) plane, respectively to create two 2D intensity images. The images are then fed into a forked CNN structure to predict the human skeletal joints. 
In \cite{an2021mars}, An {\it et al.} propose the MARS system which takes 5D radar point clouds (x, y, z, intensity and Doppler) as input and outputs human pose in several rehabilitation scenarios.
Xue {\it et al.}\cite{xue2021mmmesh} introduce mmMesh which adopts PointNet\cite{qi2017pointnet} as the feature extractor of the point cloud and incorporates SMPL\cite{loper2015SMPL} to this task, facilitating both body shape and pose predictions. 
In a follow-up work to \cite{xue2021mmmesh}, multi-subject 
3D human mesh construction is investigated~\cite{xue2022m4esh}. This is achieved by obtaining the location information from an energy map, and selectively generating 4D point clouds close to the subjects. A fine-grained human mesh is then predicted using a coarse-to-fine mesh estimation framework.
Most recently, instead of using radar point clouds, Lee {\it et al.}\cite{lee2023hupr} introduce the velocity-specific range-doppler-azimuth-elevation
map (VRDAEMap) as the input and developed a cross-modality training framework that fuses multi-scale radar features using a Cross- and Self-Attention Module (CSAM), and further refines the predicted key points through a Pose Refinement Graph Convolutional Networks (PRGCN).

The aforementioned works on mmWave-based HPE differ in the number of devices used for data collection, data representation (point clouds vs. images), and the backbone neural network architecture. 
However, none considers SUB-HPE, where there is typically limited trunk and lower limb mobility. 
A summary of the key aspects of these methods can be found in Table~\ref{table: methods comparison}.

\begin{table*}[hbt!]
% \small

\centering

\caption{
Comparison of existing works on mmWave-based HPE
}
\label{table: methods comparison}
\setlength{\tabcolsep}{0.3mm}{
\begin{threeparttable}
\begin{tabular}{l | c | c | c | c}
\toprule

Method & Radar Sensor & Ground Truth Sensor& Data Representation & Body Motions\\
\midrule

mm-Pose\cite{sengupta2020mm} & 2 TI AWR1642 & Microsoft Kinect 
&
\makecell[c]{two 2D intensity image\\(XY-plane and XZ plane)}
&
\makecell[c]{Walking Left-Arm Swing, \\Right-Arm Swing, Both-Arms-Swing}
\\

MARS\cite{an2021mars} & 1 TI IWR1443 & Microsoft Kinect & \makecell[c]{5D Point Cloud\\ (x, y, z, velocity, intensity)}
& 10 rehabilitation movements\tnote{1}

\\

mmMesh\cite{xue2021mmmesh} & 1 TI AWR1843 & VICON system &  \makecell[c]{{6D Point Cloud}\\ (x,y,z,range, velocity, energy)} 
& 8 daily activities\tnote{2}
\\
% (1) torso rotations; (2) clockwise walking; (3) counter-clockwise walking; (4) arm swing (the subject can randomly swing his/her arms horizontally or upward or downward); (5) walking back and forth; (6) walking back,and forth with arm swing; (7) walking in the place; (8) lunges (the subject keeps performing lunge pose alternatively use his/her left and right leg).

$m^4esh$\cite{xue2022m4esh} & 1 TI AWR1843 & VICON system 
& 
\makecell[c]{6D Point Cloud\\ (x, y, z, range, velocity, energy)} 
& 
\makecell[c]{7 daily activities\tnote{3} \tnote{$\dagger$} \\ freely performed by multi-person } 
% (1)walking in circles, (2)walking back and forth in straight, (3) picking up the phone from the desk, (4)putting down the phone on the desk, (5)answering phone calls while walking, (6)playing with the cell phone while sitting on the chair, (7)sitting on the chair and standing up from the chair.
\\

HuPr\cite{lee2023hupr} & 2 TI IWR1843 & RGB camera 
& \makecell[c]{VRDAEMap(velocity-specific \\range-doppler-azimuth-elevation
map)}  
& 
\makecell[c]{static actions, standing and waving \\hand(s), walking with waving hand(s)} 
\\

mmBody\cite{chen2022mmbody} & Arbe Robotics Phoenix & MoCap system 
& 
\makecell[c]{6D dense Point Cloud \\(x, y, z, velocity, amplitude, energy)}  
& 100 motions \\
\midrule

Ours & 2 TI IWR6843 & OptiTrack system 
& 
\makecell[c]{intensity point cloud, \\ Doppler point cloud} 
& \makecell[c]{upper limb movements,  head rotation, \\ driving simulation}\\

\bottomrule
\end{tabular}
\begin{tablenotes}   
\scriptsize        
\item[1] Right/left/both limb extension, right/left side lunge right/left front lunge, right/left upper body extension, squad.
\item[2] Torso rotations, clockwise walking, counter-clockwise walking, arm swing), walking back and forth; walking back, and forth with arm swing, walking in the place, lunges.
\item[3] Walking in circles, walking back and forth in straight, picking up the phone from the desk, putting down the phone on the desk, answering phone calls while walking, playing with the cell phone while sitting on the chair, sitting on the chair and standing up from the chair.
\item[$\dagger$] Freely performed by multi-person in one recording. 
\end{tablenotes}
\end{threeparttable}
}
\end{table*}

\subsection{Public mmWave-based HPE datasets}
Very few public datasets are currently available for mmWave-based HPE.  
In \cite{an2021mars}, the authors release a dataset MARS containing radar point clouds and annotation obtained using Microsoft Kinect V2 sensor.
Chen {\it et al.} proposed mmBody\cite{chen2022mmbody}, a multi-scenario RGBD-paired mmWave radar (Arbe Robotics Phoenix) point cloud dataset for human pose reconstruction with 3D ground truth provided by a motion capture system. 
The work in \cite{lee2023hupr} also provides a dataset HuPR, which contains {raw radar data together with 3D annotation generated from a synced RGB camera. 

With the exception of HuPR, the aforementioned public datasets only contain intermediate representations of the radar data, e.g., point clouds.} The lack of raw data greatly limits innovations on radar signal processing algorithms and consequently affects the informativeness of training data to the HPE models. Another limitation of some datasets (e.g., HuPR and MARS) lies in the absence of accurate ground truth due to the use of RGB or RGB-D inputs for annotations.

\section{Methodology}
\label{sect:method}
SUPER considers the problem of estimating upper body human poses when a subject faces mmWave radar sensors at a known distance.  The assumption for known distance is valid in confined environments such as in an office cubicle or inside a vehicle. Alternatively, existing approaches for mmWave-based target localization can be adopted to determine a bounding box around the subject~\cite{chen23mmTracking}.
In this section, we first provide an overview and the design rationale of the SUPER pipeline and then present details of its individual components.

\subsection{Overview and Design Rationale}
\begin{figure}[h]
  \centering
  %\includegraphics[width=0.45\textwidth, height=0.2\textheight]{images/systemdiagram.jpg}
  \includegraphics[width=\linewidth]{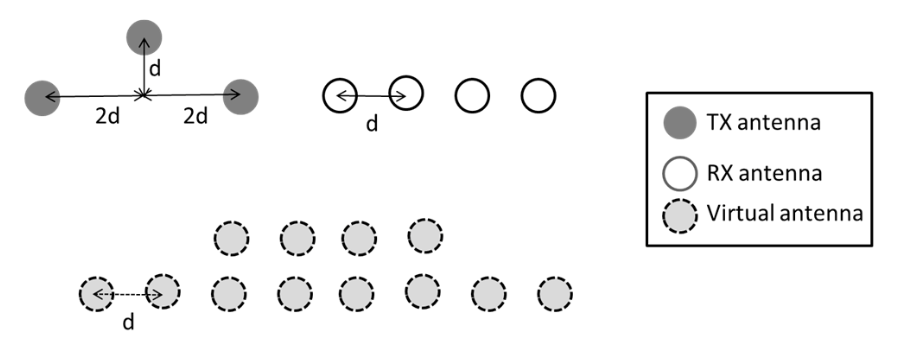}
  \caption{A 2-Dimensional MIMO antenna array for IWR6843ISK radar. The separation $d$ equals half wavelength.}
  \label{fig:virtualantenna}
\end{figure}

Low-end COTS mmWave radars typically have a small number of Tx and Rx antennas, which restrict their spatial resolution. Take TI IWR6843ISK radar as an example. It features 3 Tx and 4 Rx antennas forming a 12 virtual antenna array as illustrated in Figure~\ref{fig:virtualantenna}. 
Placed horizontally, this configuration results in angle resolutions of 15 degrees and 55 degrees, respectively, in the horizontal and vertical directions. To estimate fine-grained SUB-HPE, a high azimuth angle resolution is necessary for extremities when the arms are extended while a high elevation angle resolution is helpful in distinguishing subtle head and trunk poses. To mitigate the limitations of low-end mmWave radars, we employ two closely located radar sensors: one oriented horizontally and the other vertically. Despite the lack of coordination, the reflected wave from one radar's transmission is unlikely mistaken as that from the other radar since the resulting range bins are outside the region of interest (ROI). Note that although dual-radar systems have been also employed in mm-Pose~\cite{sengupta2020mm} and HuPR~\cite{lee2023hupr}, the data is used to produce 2D heatmaps (images) in perpendicular planes rather than being fused together in 3D point clouds. 

Several existing mmWave-based HPE methods model human body as a point cloud, which is obtained from range-Doppler maps over multiple chirps of radar signals. Doppler information has sufficient coverage on the entire body only if there are significant motions in different body parts. In seated positions, however, movements in the trunk and low limbs are confined leading to sparse points in space. In contrast, the intensity of reflected signals from the bulk of the body is high regardless of motions as long as the subject is sufficiently close to the radars. Thus, a range-angle map, augmented with intensity information from a multi-antenna system, better captures the occupation of human body in space. Motivated by this observation and with the unique characteristics of seated SUB-HPE in mind, we extract two point clouds with reflected intensity and Doppler information. The ablation study in Section~\ref{sect:eval} further substantiates the empirical evidence supporting the complementary nature of the two input sources.

%From each radar sensor, we extract both intensity and Doppler information to extract 3D human poses and motions. Section [ref] contains an ablation study validating our assertions and elucidating the contributions of these components to system performance.
%
%To the best of our knowledge, previous radar sensor work has focused on generating point clouds exclusively based on range-Doppler maps. These studies have primarily involved subjects engaged in physical activities with significant body movements. However, in scenarios where subjects remain stationary or exhibit minimal movement, such as drivers constrained by seat belts while operating a vehicle, all radar reflections from the human body in the same range, collapse into a single point on the range-Doppler map, rendering it inadequate for describing the entire human body. Conversely, the range-angle map, derived directly from signals reflected across the entire human body, depends on the range and the reflectivity of human skin, rather than the subject's movements. This range-angle map, augmented with intensity information from a multi-antenna system, captures the spatial occupation of the human body. This is especially valuable in situations involving hybrid or immobile subjects.

\begin{figure*}[tb]
  \centering
  %\includegraphics[width=0.8\textwidth, height=0.2\textheight]{images/systemdiagram.jpg}
  \includegraphics[width=0.8\linewidth]{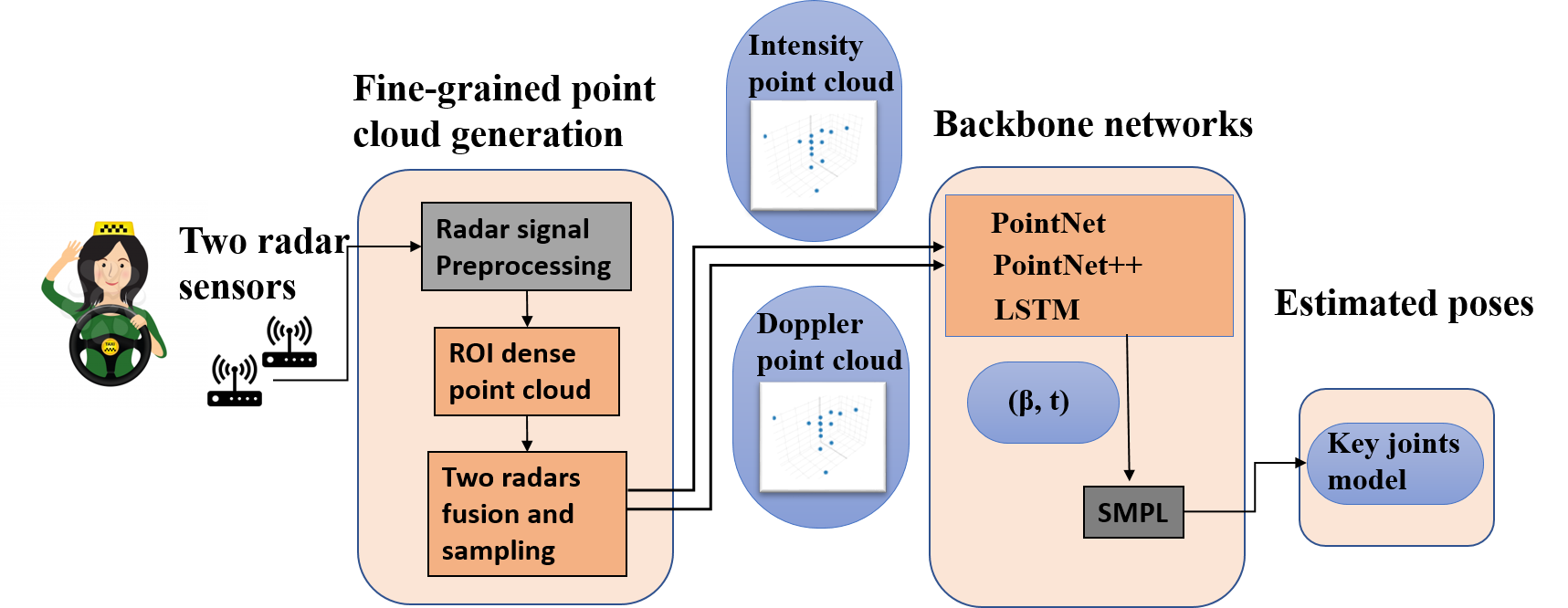}
  \caption{The system diagram of SUPER. New processing blocks introduced in this paper are highlighted in orange, and intermediate data flows are highlighted in blue.}
  \label{fig:systemdiagram}
\end{figure*}

The overall system diagram of SUPER shown in Figure~\ref{fig:systemdiagram}, consists of two main processing blocks, i.e., point cloud generation and a backbone network. The reflected RF signals from two radar sensors are preprocessed using match filtering and range-FFT. Dense point clouds are then generated by sampling the ROIs in 3D space centred around each radar. A dual-radar fusion algorithm coherently combines data from two radars and samples the results to produce fine-grained point cloud data representation for intensity and for Doppler. Both point clouds are fed into the backbone network. The network comprises building blocks from PointNet~\cite{charles2017pointnet}, PointNet++~\cite{qi2017pointnet++}, and LSTM to extract global and local features to predict the SMPL pose parameters in each frame. The pipeline can be easily extended to predict body shape parameters and will be investigated as part of future work.

\subsection{Point cloud generation with dual-radar fusion}

% The pipeline from raw radar data to dense pc
\begin{figure}[tb]
  \centering
  %\includegraphics[width=0.45\textwidth, height=0.2\textheight]{images/systemdiagram.jpg}
  \includegraphics[width=\linewidth]{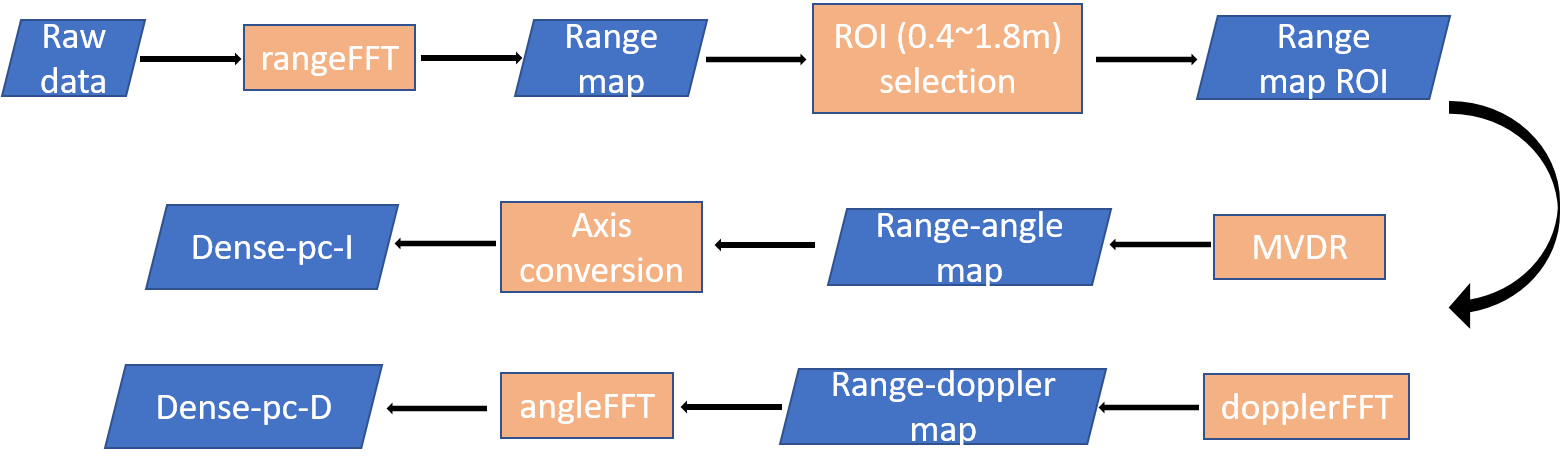}
  \caption{Generation of dense point clouds from raw radar data. One intensity point cloud and one Doppler point cloud are produced for each radar separately.}
  \label{fig:raw2densepc}
  % \Description{The system diagram of the proposed processing chain to generate dense point cloud from raw radar data. From left to right, they are the radar sensor, the chest-wall movement estimation block, the vital sign estimation block and the vital sign output block.}
\end{figure}

In this section, we introduce a novel pipeline to generate quality point clouds from data collected by two closely located radars. Data from each radar goes through separate branches to handle intensity and Doppler information. The overall processing consists of two stages: the first stage transforms raw radar data to a dense point cloud, which acts as an intermediate representation. In the second stage, data from the two radar sensors are fused together and then sampled to produce a fine-grained point cloud.

\subsubsection{Dense point cloud generation}
Raw I-Q samples from each radar in intermediate frequency (IF) follow the standard pre-processing steps. These include mapping the raw radar data into a range map through range-FFT and DC compensation to eliminate static background clutters.
As previously mentioned, SUPER operates under the assumption that the approximate distance between the subject and the radars is known. This knowledge enables the designation of an ROI that encompasses the subject. For example, when seated around 1 meter away from the radars, the range bins that span the subject's body are approximately from 0.4 meters to 1.8 meters. These parameters can be easily adjusted given the setup of different scenarios.

\paragraph*{Intensity point clouds} To generate intensity point clouds, we further consider 180-degree field of view (FOV) in both horizontal and vertical directions and choose a non-uniform sampling scheme as shown in Table~\ref{tab:anglesampling}. Specifically, for the radar placed horizontally (radar H), $\theta$ and $\phi$ correspond to the azimuth and elevation angles; for radar V, the reverse is true. Clearly, as indicated in Table~\ref{tab:anglesampling}, angles are densely sampled in the axis where more spaced virtual antennas are available and near the center, while in the perpendicular direction, fewer angle bins are sampled. Consequently, amongst the 30 range bins between 0.4 meters and 1.8 meters from the subject, there are in total 6930 ($= 21\times 11\times 30$) sample points in the ROI. 

\begin{table}
%\resizebox{\textwidth}{!}
% \small
\footnotesize
  \begin{center}
    \caption{Non-uniform Angle Sampling (unit in degree) }
    \label{tab:anglesampling}
    \begin{tabular}{c |c c c c c c c c c c c}
    \toprule
 %	   \multicolumn{8}{c}{(top) maximum correlations } \\
 %	   \multicolumn{8}{c}{(bottom) corresponding delays (ms)} \\
 %	\midrule
 		$\theta$ & -70 & -60 & -50 & -40 & -30 & -25 & -20 & -15 & -10 & -5 & 0 \\
                    & 5 & 10 & 15 & 20 & 25 & 30 & 40 & 50 & 60 & 70 \\
 		% \hline
   \midrule
 		$\phi$ & -70 & -50 & -30 & -20 & -10 & 0 & 10 & 20 & 30 & 50 & 70 \\
 	\bottomrule
    \end{tabular}
  \end{center}
\end{table}

Next, we apply Minimum Variance Distortionless Response (MVDR) to generate an intensity spectrum for each point location in the ROI. We first estimate the correlation matrix for each range index $i$, using all $N$ chirps within one frame,
\begin{align*}
R_i &= \frac{\sum_{n=1}^N \mathbf{y}\mathbf{y}^H }{N}, \\
R_i &= R_i + \alpha \frac{trace(R_i)}{K} I_K,
\end{align*}
where $\mathbf{y}$ is a column vector of the received signal at each antenna, $N$ is the number of chirps in one frame, $K$ is the number of received antennas, and $\alpha$ is a control parameter to prevent singularity.

Next, we calculate the steering vector $\mathbf{a_s}$ from the virtual antennas array as
\begin{align*}
\mathbf{a_s}(n)= 
\left\{
  \begin{array}{lr} 
      exp(j\pi (\mu_a (n-1))), & 1 \leq n\leq 8, \\
      exp(j\pi (\mu_a (n-6-1) + \mu_b)), & 9 \leq x \leq 12, 
    \end{array}
\right.
\end{align*}
where
\begin{align*}
\mu_a &= sin(\theta \pi/180) cos(\phi \pi/180), \\
\mu_b &= sin(\phi \pi/180).
\end{align*}

Finally, we calculate the intensity spectrum for each sample point as
\begin{align*}
IS(\theta,\phi,i) = \frac{1}{\mathbf{a_s}^H R_i^{-1}\mathbf{a_s}},
\end{align*}
where $\mathbf{a_s}$ is the steering vector, and $i$ is the range index. This process creates a 4D point cloud with intensity values in polar coordinates, which can then be transformed into a dense point cloud in a Cartesian coordinate system centered on a radar. 

\paragraph*{Doppler point clouds} To generate Doppler point clouds, we follow a similar procedure to that in \cite{xue2021mmmesh}. Specifically, Doppler-FFT on the chirps in a frame is applied to derive 2D range-Doppler maps ($30\times 128$) of each received antenna. For every point in the 2D range-Doppler map, its velocity and power are calculated through an additional angle-FFT across multiple received antennas. The procedure is applied to data from the two radars independently, resulting two 5D point clouds of 3840 ($= 128 \times 30$) points for each radar. 
%This processing chain is the most popular in literature, and more details can be found in reference~\cite{xue2021mmmesh}. It is essential to note that the horizontal radar sensor serves as the origin of our coordinate system, as depicted in Figure~\ref{fig:radars}. Accordingly, we shift all the coordinate data from the vertical radar to this origin based on the actual positions of the two radar sensors.

It is worth noting that the term ``dense" is adopted to differentiate this representation from the eventual fused point clouds. While the point clouds in this initial stage remain relatively sparse when compared to those generated by Lidar sensors, they are denser than the point clouds typically found in existing literature on mmWave-based HPE. This increased density is achieved through spatial oversampling in the intensity point clouds. Further information regarding the process is illustrated in Figure~\ref{fig:raw2densepc}.

\subsubsection{Dual-radar fusion for fine-grained point clouds}
% The pipeline from dense pc to final pc
\begin{figure}[tb]
  \centering
  %\includegraphics[width=0.45\textwidth, height=0.2\textheight]{images/systemdiagram.jpg}
  \includegraphics[width=\linewidth]{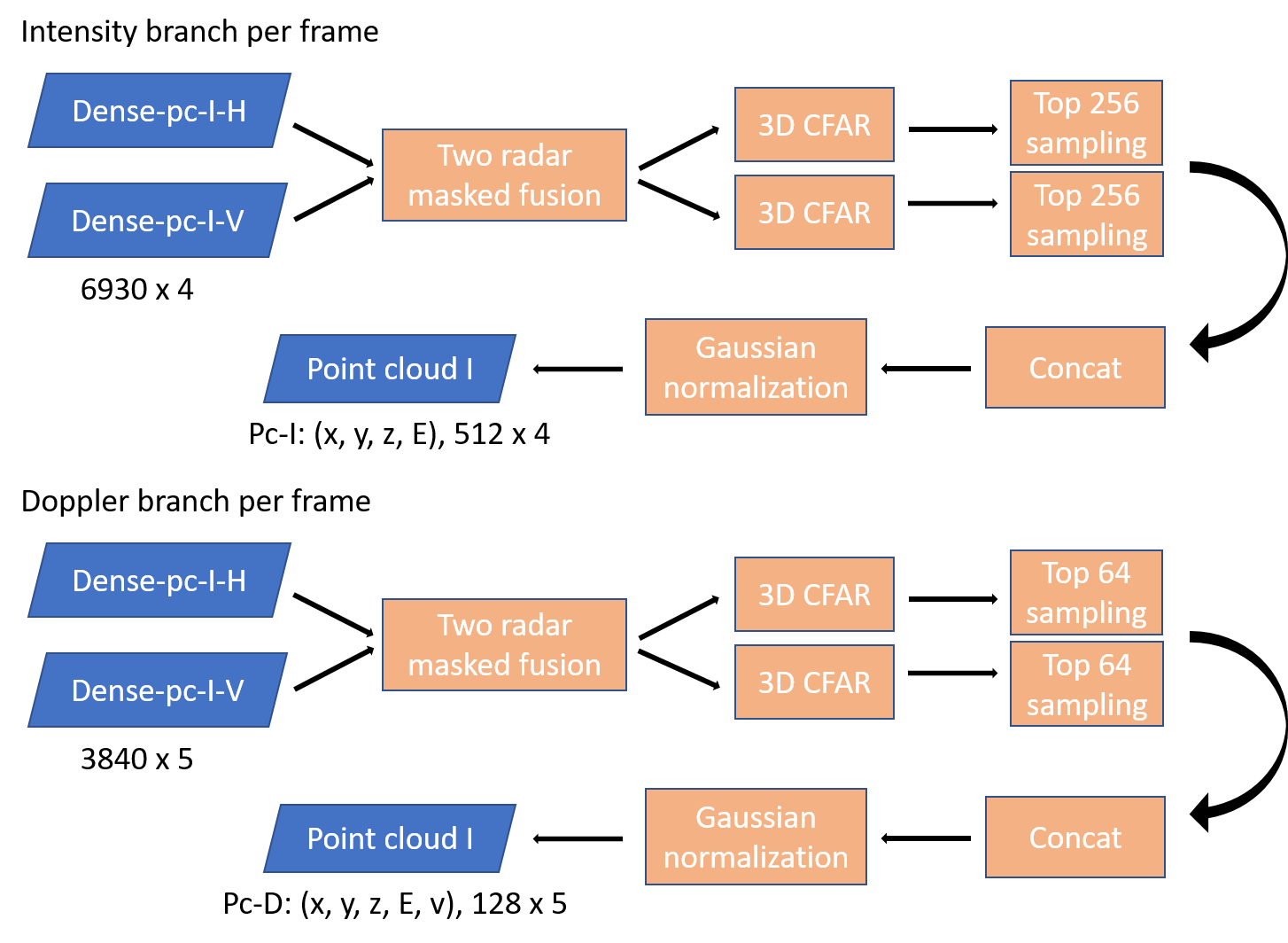}
  \caption{Generation of fine-grained point clouds by fusing and sampling dense point clouds from the two radars.}
  \label{fig:dense2pc}
  % \Description{The system diagram of the proposed processing chain to generate dense point cloud from raw radar data. From left to right, they are the radar sensor, the chest-wall movement estimation block, the vital sign estimation block and the vital sign output block.}
\end{figure}
To this end, we have generated four point clouds, i.e., one 4D intensity point cloud and one 5D Doppler point cloud from each radar. 
The two radar sensors are positioned in close proximity, approximately 15cm apart. Thus, the dense point clouds generated by each radar sensor roughly share the same ROI but are complementary spatially. Radar H captures detailed information in the horizontal direction, which can be used to enhance the quality of the point cloud derived from radar V, and vice versa. Therefore, the purpose of dual-radar fusion is two-folded. First, it refines the point clouds from one radar using the point clouds from the other radar.  Second, it trims the over-sampled point clouds and retains only salient points. At the end of the procedure, a single intensity point cloud and a single Doppler point cloud are obtained for further processing. An overview of this process is given in Figure~\ref{fig:dense2pc}.

% \begin{figure}[tb]
%   \centering
%   % Requires \usepackage{graphicx}
%   \includegraphics[width=\linewidth]{images/maskfusion2.png}
%   \caption{Two Radars Masked Fusion Algorithm.}
%   \label{fig:maskfusion}
% \end{figure}

\paragraph*{Masked refinement} To refine the point clouds from both radars, we first transform their representations from polar coordinate frames to a unified Cartesian coordinate frame. Consider the 4D intensity point clouds from radar H as an example. A similar procedure is applied to the intensity point cloud from radar V and the 5D Doppler point clouds from both radars. Let the point cloud from radar H be the target and that from radar V serves as a reference. For each point in the target point cloud, the $K$ nearest points in the reference point cloud are identified.
The mean power value of these points is computed through averaging. The value of the point in the target point cloud is replaced by the product of itself and the mean value. This multiplication has the effect of masking or suppressing points with high values in only one point cloud and amplifying those with high values in both. Furthermore, the operation can preserve local power variations, as the masks within the same local area are nearly identical.

% \begin{figure}[tb]
%   \centering
%   % Requires \usepackage{graphicx}
%   \includegraphics[width=\linewidth]{images/3dcfar2.png}
%   \caption{3D CFAR Algorithm.}
%   \label{fig:3dcfar}
% \end{figure}
\paragraph*{Point cloud trimming} Due to spatial over-sampling, the dense point clouds produced thus far contain redundant information. To retain only informative points, we extend the principles of the {2D Constant False Alarm Rate (CFAR) algorithm~\cite{mark2005RSP} and } implement a 3D CFAR algorithm, by adaptively calculating thresholds to detect local peaks as key points. Finally, we output the top 256 key points from the intensity point clouds and the top 64 key points for the Doppler point clouds.

Following the extraction of key points, we merge the point clouds from both radars and apply a Gaussian normalization filter to the values. The final fine-grained point cloud consists of 512 key points, featuring $[x, y, z, intensity]$ for intensity, and 128 key points with attributes $[x, y, z, power, velocity]$ for Doppler. An example fine-grained point clouds generated from the process is shown in Figure~\ref{fig:examplepcs}. { In this example, the subject raises their right hand to the top. It is evident from Figure~\ref{fig:intensitypc}, the intensity points are present not only around the raised arm but also at other areas of the upper body. In contrast, as shown in Figure~\ref{fig:dopplerpc},
the Doppler points mainly appear around the raising arm with non-negligible velocity. }
\begin{figure}[tp]
    \centering
    \begin{subfigure}[b]{0.49\columnwidth}
        \centering
        \includegraphics[width=1.0\textwidth, height=0.8\textwidth]{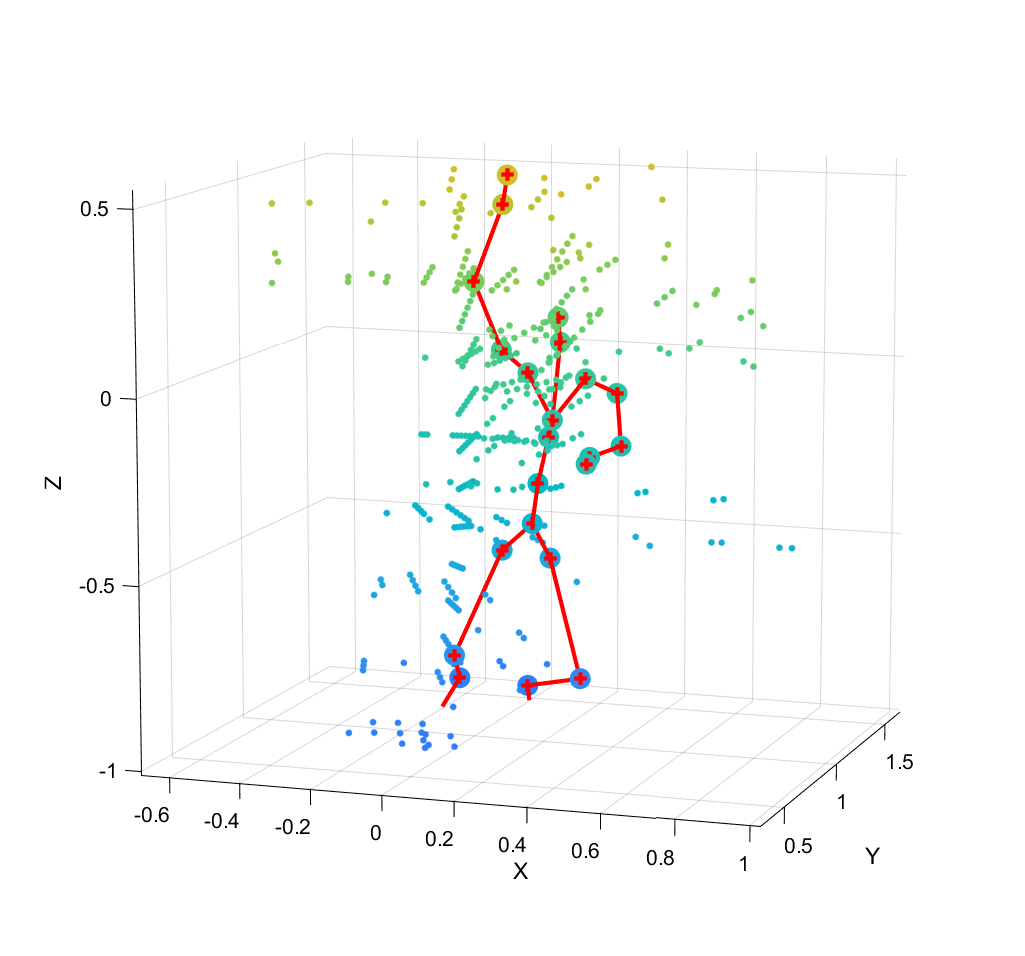}
        \caption{An intensity point cloud.}
        \label{fig:intensitypc}
    \end{subfigure}
    %\hfill
    \begin{subfigure}[b]{0.49\columnwidth}
        \centering
        \includegraphics[width=1.0\textwidth, height=0.8\textwidth]{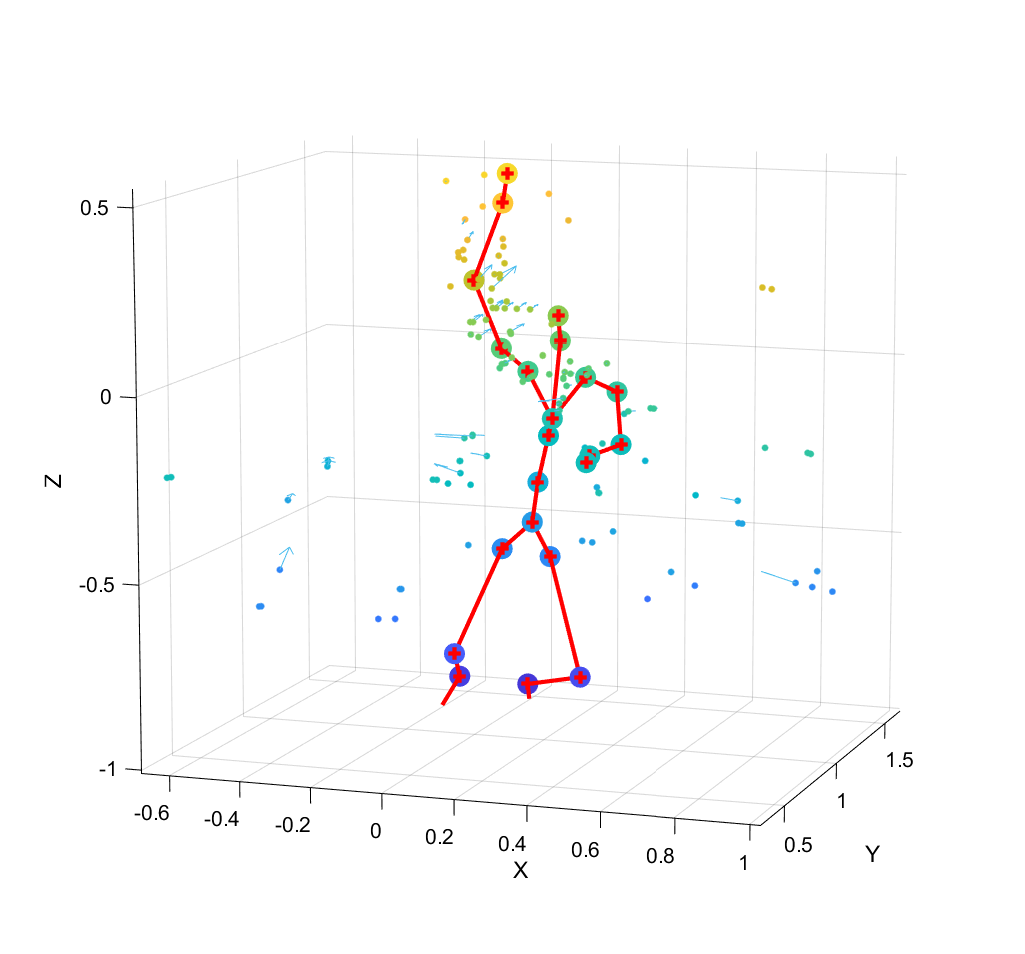}
        \caption{A Doppler point cloud.}
        \label{fig:dopplerpc}
    \end{subfigure}
\caption{{An example fine-grained point clouds. Ground truth skeleton is shown in red. The magnitude and direction of Doppler velocity are shown in arrows}}
\label{fig:examplepcs}
\end{figure}

\subsection{The deep neural network backbone}
% The DNN architecture
\begin{figure}[tb]
  \centering
  %\includegraphics[width=0.45\textwidth, height=0.2\textheight]{images/systemdiagram.jpg}
  \includegraphics[width=\linewidth]{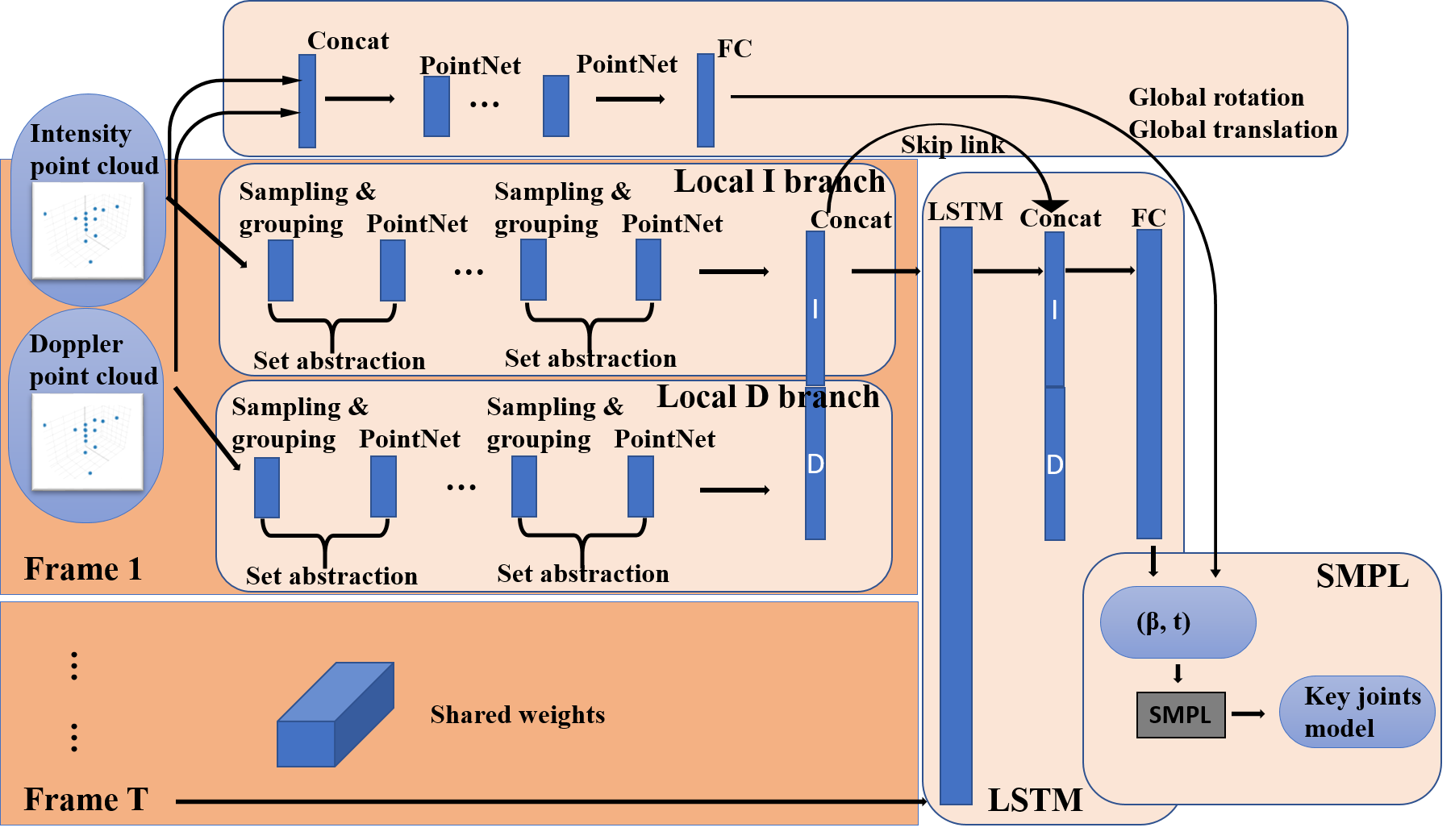}
  \caption{The architecture of the deep neural network backbone.}
  \label{fig:dnn_architecture}
  % \Description{The system diagram of the proposed processing chain to generate dense point cloud from raw radar data. From left to right, they are the radar sensor, the chest-wall movement estimation block, the vital sign estimation block and the vital sign output block.}
\end{figure}

A deep neural network (Figure~\ref{fig:dnn_architecture}) is designed to take multiple frames of fine-grained point clouds as inputs to predict joint positions in a human skeleton model. 
The network incorporates both global and local contexts to estimate the intricate translation and rotation dynamics. To capture the global context, we include a dedicated branch that stacks three basic PointNet blocks~\cite{charles2017pointnet}. To extract local information, three hierarchy set abstraction layers in PointNet++ are stacked to process both the intensity and Doppler point clouds~\cite{qi2017pointnet++}.

Furthermore, to exploit the temporal dependencies between frames, two layers of unidirectional Long Short-Term Memory (LSTM) cells are used~\cite{hochreiter1997LSTM}, spanning $T=20$ steps or frames (equivalent to one second). To enhance information flow, a skip/residual link is introduced that connects features prior to the LSTM layers and post-LSTM. Finally, after several fully connected (FC) layers, the model outputs rotations of each joint within the human skeleton model. To improve the accuracy of rotation estimation, following\cite{zhou20196Drotation}, we represent joint rotations using 6D parameters of the rotation matrices rather than 3D axis angles.

The model subsequently leverages SMPL to generate the final joint positions. A gender-neutral model is used by fixing the default shape parameters. For seated upper body poses, we freeze the rotation parameters of joints in the lower body and only estimate the positions of the upper body joints (14 joints)~\cite{loper2015SMPL}.

The loss function is defined as the mean square error (MSE) of the joint coordinates: 
\begin{align}
Loss = \frac{1}{F} \sum_{f=1}^{F} ||P_{f,J}^{(f)} - P_{gt,J}^{(f)}||_2,
\end{align}
where $f$ is the frame index, $F$ is the total number of frames in the batch, $J$ denotes the joint set,  $P_{f,J}^{(f)}$ is the estimated positions of key joints, and $P_{gt,J}^{(f)}$ is the corresponding ground truth positions. Note that the loss is a function of the pose parameters ($\beta$) and global translation ($t$). From the experiments, we find that instead of directly regressing the joint positions, passing the joint rotation parameters through SMPL to estimate the resulting joint position errors results in higher accuracy and faster convergence. This can be interpreted as a non-linear transformation of the MSE loss function using the SMPL model. 

The total number of learning parameters in the network is 2.9 million or {2.65G FLOPs}. Incoming point clouds are processed in a sliding window manner with a window size of 20 frames. 

\section{Implementation and Datasets}
\label{sect:impl}
In this section, we present the implementation of a prototype SUPER system using COTS mmWave radars and the experiment results from multi-subject testbed evaluations under various conditions, which are purposely chosen to closely mimic real-life situations.
\subsection{Implementation}
\label{subsec:experimentsetups}

\begin{figure}[th]
    \centering
    \begin{subfigure}[b]{0.7\columnwidth}
        \centering
        \includegraphics[width=\textwidth]{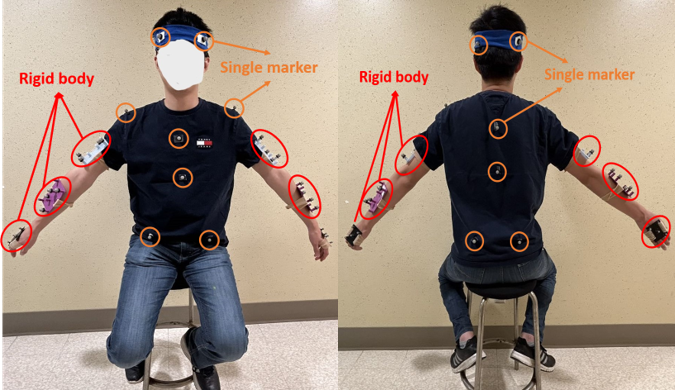}
        \caption{Markers placement: front and back.}
        \label{fig:markers}
    \end{subfigure}
    %\hfill
    \begin{subfigure}[b]{0.7\columnwidth}
        \centering
        \includegraphics[width=\textwidth]{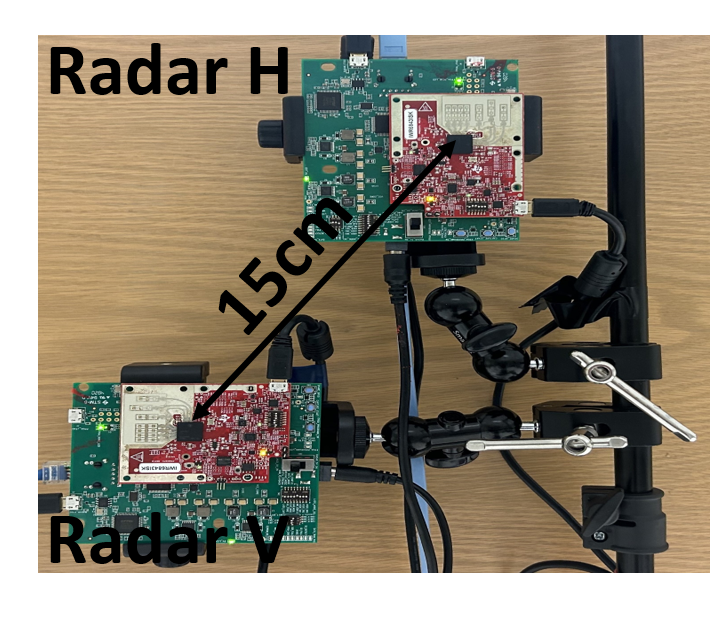}
        \caption{Co-located radar sensors.}
        \label{fig:radars}
    \end{subfigure}
\caption{Experiment setup: markers and radars.}
\label{fig:markersandradars}
\end{figure}

Two IWR6843ISK boards~\cite{iwr6843ISK} together with DCA1000EVM boards~\cite{DCA1000EVM} are used in the experiments. The radar boards operate at $60 \sim 64$ GHz (with 4-GHz bandwidth) and transmit FMCW signals. The radar front-ends include $3$ transmit antennas (Tx), $4$ receive antennas (Rx), with $120^{\circ}$ azimuth field of view (FoV) and elevation FoV. The 3 transmitting antennas emit FMCW chirps in a time-division manner, which results in a 12 virtual antennas array. Each FMCW chirp is composed of 225 sampling points, and the frequency of RF will increase from 60 GHz to 64 GHz. 128 chirps constitute one frame at a frame rate of 20Hz. The acquired raw IF signal is sent to a host PC via Ethernet, where mmWave Studio~\cite{mmwavestudio} is used to initiate, configure, and control the radar boards. The detailed radar sensor settings is summarized in Table~\ref{tab:radarsettings}.

\begin{table}
%\resizebox{\textwidth}{!}
\small
% \footnotesize
  \begin{center}
    \caption{Radar Hardware Settings.}
    \label{tab:radarsettings}
    \begin{tabular}{c | c | c }
    \toprule
 %	   \multicolumn{8}{c}{(top) maximum correlations } \\
            parameters & description & values \\
    \midrule
 		$N_{tx}$ & number of transmit antennas & 3  \\
            $N_{rx}$ & number of receive antennas & 4  \\
            $N_{virtual}$ & number of virtual antennas & 12  \\
            $P_{f}$ & frame duration & 50 (ms)  \\
            $f_{s}$ & start frequency & 60 (GHz)  \\
            $f_{e}$ & end frequency & 64 (GHz)  \\
            $t_{rs}$ & start ramp time & 0 ($\mu$s)  \\
            $t_{re}$ & end ramp time & 58 ($\mu$s)  \\
            $t_{idle}$ & chirp idle time & 7 ($\mu$s)  \\
            $N_{adc}$ & number of samples per chirp & 225  \\
            $N_{chirp}$ & number of chirps per frame & 128  \\
    \bottomrule
    \end{tabular}
  \end{center}
\end{table}

The preprocessing steps and point cloud generation are implemented in MATLAB R2021a, which takes raw IF signals as input, and outputs the fine-grained 3D point cloud data. The neural network backbone is implemented in PyTorch.

\subsection{Data collection procedure}
To evaluate SUPER’s performance, we recruited 10 participants (3 females and 7 males), aged between 21 and 46, and with BMI in the range of $18.1 \sim 31.6$. Participants wore their daily attire such as T-shirts, blouses, and sweaters of different fabric materials. 
%9 participants have extensive 3+ years driving experience and 1 participant has 1 year driving experience. 
This research protocol has been approved by the research ethical board (REB) from our institution.

Both radar and mocap data are collected in a $6.5m \times 6m$ lab. The lab (Figure~\ref{fig:labsetup}) has standard office furniture and many electronic equipment and wireless transceivers (WiFi, LTE, Bluetooth, etc.). Both radar sensors on a tripod as in Figure~\ref{fig:radars} with 1.5 meters high and 1 meter away from the subjects and oriented at a 20-degree horizontal angle. We define a local coordinate system with respect to radar H. During the experiments, only one subject is present in the predefined position.

Ground truth of subject poses are collected from  OptiTrack, a motion capture system~\cite{optitrack} with 12 cameras. Both radar sensors and the OptiTrack system are synchronized after data collections at frame level using ``synchronization" motions at the beginning of each trial. 
The output of the OptiTrack system are coordinates of markers and rigid bodies on the body of participants as shown in Figure~\ref{fig:markers}. We utilize MotionBuilder~\cite{motionbuilder} to build a customized human actor for each participant and generate accurate joints coordinates through motion tracking functionalities built in the software. Videos have been recorded during data collection for reviewing purposes but are not further processed. 

\begin{figure}[th]
  \centering
  %\includegraphics[width=0.45\textwidth, height=0.2\textheight]{images/systemdiagram.jpg}
  \includegraphics[width=0.8\linewidth]{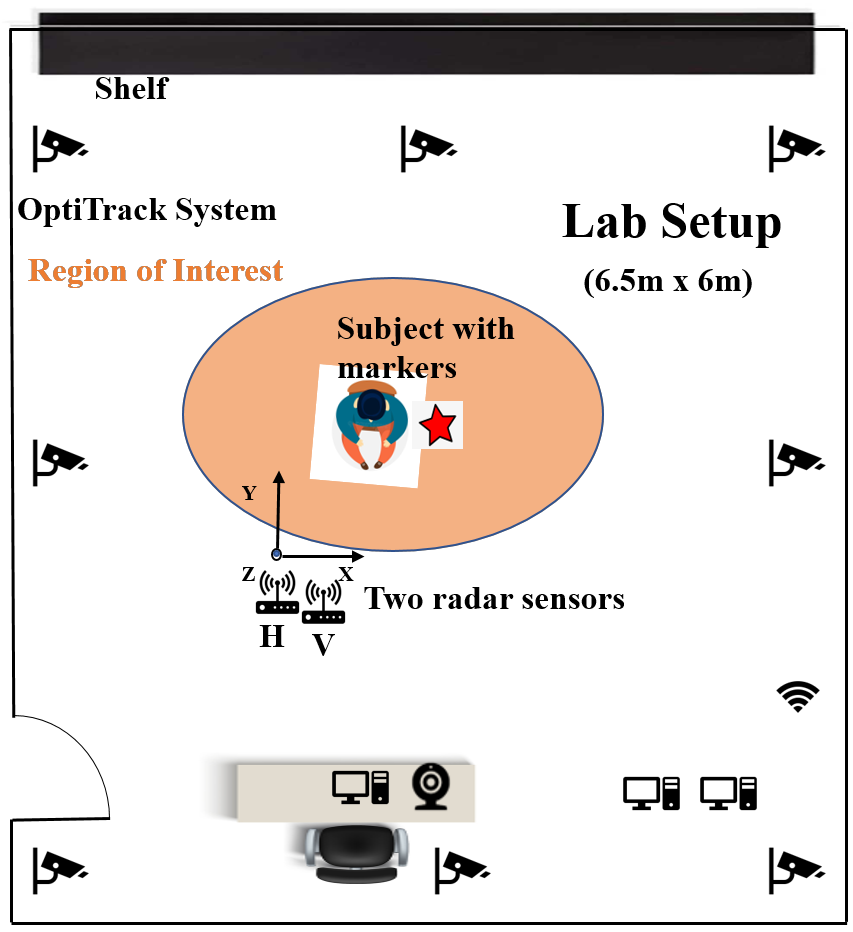}
  \caption{The Lab environment for data collection.}
  \label{fig:labsetup}
\end{figure}

% \begin{figure}[th]
%     \centering
%     \begin{subfigure}[b]{0.49\columnwidth}
%         \centering
%         \includegraphics[height=0.35\textheight]{images/labsetup2-1.png}
%         \caption{Lab}
%         \label{fig:setup1}
%     \end{subfigure}
%     %\hfill
%     \begin{subfigure}[b]{0.49\columnwidth}
%         \centering
%         \includegraphics[width=\linewidth]{images/loungesetup2.jpg}
%         \caption{Lounge room}
%         \label{fig:setup2}
%     \end{subfigure}
% \caption{Two environments for data collection.}
% \label{fig:setup}
% % \Description{The main Lab environment and the real-life lounge room for radar sensing experiments.}
% \end{figure}

During the data collection process in the controlled laboratory environment, subjects engaged in three distinct motion sequences that are designed to mimic movements while seated in confined environments. These include: hand-reaching, driving, and head rotation.
A Microsoft Xbox Gaming steering wheel is used to mimic a driving platform and is placed in front of the subjects.

\begin{itemize}
\item{\it Hand-reaching trials:} Participants were instructed to use their right hand to interact with hypothetical objects in their surroundings while keeping their left hand stationary. These trials included interacting with objects positioned directly above one's head (top), in the top front (up-front), in front but to the side (right-front), to the side (right), and below (bottom). 
%This trial was meticulously designed to simulate the multifaceted activities and interactions that occur between drivers and modern smart vehicles, reflecting scenarios where drivers must seamlessly switch between various in-car controls and driving tasks.

\item{\it Driving trials:} These trials aimed to replicate common driving activities. Subjects were instructed to perform routine driving (with both hands on the wheel), conduct traffic checks (by leaning forward and inspecting both left and right directions), engage in a conversation with a passenger (rotating the head towards the passenger), execute reverse maneuvers (turning the head to see one's back over the shoulder), and operate the control panel (reaching the right-front area and virtually press buttons with one's right hand). 

\item{\it Head rotation trials:}
These trials capture deliberate head movements while keeping one's torso mostly stationary. Subjects were instructed to look left and right, up and down, and upper/lower left/right, etc. 
\end{itemize}

\subsection{The dataset}
In total, we conducted 30 trials from 10 participants, with each lasting around 10 minutes. The total number of radar frames collected is around 360,000 from each radar sensor. The total size of all raw radar data in the dataset is around 900GB. {The ground truth data for each frame contains joint angles and positions of 14 upper body key joints\footnote{The local body joints are set to fixed sitting poses} and the global translations. The total size of the ground truth data is around 900MB.
The dataset is organized by subject ID (de-identified), trial name, and data types (radar data vs ground truth data). }

\section{Performance Evaluation}
\label{sect:eval}
In this section, we present the performance of SUPER and ablation studies. 
\subsection{Evaluation metrics and baseline method}
We chose three metrics in literature to quantify the accuracy of the estimated upper body joints. The first one is the Mean Per Joint Position Error (MPJPE), which measures the {\it absolute} average distance (mm) between the predicted joints of a human skeleton and the ground truth joints in a given dataset. The MPJPE is defined as:
\begin{align}
E_{MPJPE} (f, J) = \frac{1}{K_{J}} \sum_{k=1}^{K} ||P_{f,J}^{(f)}(k) - P_{gt,J}^{(f)}(k)||_2,
\end{align}
where $f$ denotes a frame, $J$ denotes the joints model/set, $K$ is the number of joints in the model/set, $P_{f,J}^{(f)}(k)$ is the estimated position of joint $k$, and $P_{gt,J}^{(f)}(k)$ is the corresponding ground truth position. Finally, the MPJPEs are averaged over all frames.

The second metric is the Procrustes alignment MPJPE (PA-MPJPE) that calculates the average 3D joint distance (mm) after performing Procrustes alignment~\cite{gower1975PA} on the estimated and ground-truth joint sets. PA-MPJPE measures how well the pose estimation model captures the structural information of the pose, rather than just its location or scale. It eliminates system biases and allows for fair comparisons across different scales of the same pose.

The third metric is the percentage of correct keypoints under a distance threshold e.g. $15mm$ ($PCK@15mm$). This metric is defined as:
\begin{align}
PCK@15mm = \frac{1}{K} \sum_{k=1}^{K} \delta_k,
\end{align}
where $K$ is the total number of keypoints (joints), $\delta_k$ is a binary value indicating whether the distance between the ground truth keypoint and the predicted keypoint is within a certain threshold.

\paragraph*{Baseline method}
%We chose to use the solution in mmMesh~\cite{xue2021mmmesh} as our benchmark.This choice sizzles with originality because, surprisingly, no one has previously delved into upper body pose estimation in hybrid or immobile scenarios like ours. Among others, mmMesh is one of the most relevant and up-to-date methods in this field. Consequently, we adopted mmMesh as our benchmark for comparison.
We adopt mmMesh~\cite{xue2021mmmesh} as the baseline method for comparison. The choice is primarily driven by the fact that the model architecture was made publicly available by the authors. Although $m^4esh$ is more recent, it targets multi-subject scenarios, which are outside the scope of this paper. 
The mmMesh model parameters were retrained using the hyperparameters suggested in \cite{xue2021mmmesh} and data from radar H only for consistency. 
%to compare with. This choice sizzles with originality because, surprisingly, no previous work delved into upper body pose estimation in hybrid or immobile scenarios like ours. To the best of our knowledge, mmMesh is the most relevant and up-to-date method in this field and is the only mmWave radar point cloud based work that provides their implementation of radar preprocessing part. 

\subsection{Main results}
\begin{table}[!t]
% \small

\centering

\caption{
Accuracy of joint estimations (in mm)
}
\label{table:results_main}
\begin{adjustbox}{width=\linewidth}
\begin{tabular}{l | l |c | c | c }
\toprule

action & method
& {MPJPE$\downarrow$} 
& {PA\_MPJPE$\downarrow$}
& {PCK@15mm$\uparrow$}
\\

\midrule

\multirow{2}*{driving}

& mmMesh    
& 156.85$\pm$25.18 & 29.60$\pm$6.2 & 13.76$\pm$7.38
\\

& \textbf{Ours}
& \textbf{112.46$\pm$12.70} & \textbf{16.32$\pm$2.45} & \textbf{27.38$\pm$11.15} 
\\
\midrule

\multirow{2}*{handreaching}

& mmMesh
& 148.33$\pm$25.18 & 26.97$\pm$4.39 & 15.74$\pm$8.65
\\

& \textbf{Ours}
& \textbf{114.87$\pm$25.07} & \textbf{15.19$\pm$2.56} & \textbf{37.17$\pm$8.28}
\\
\midrule

\multirow{2}*{head rot.}

& mmMesh
& 174.42$\pm$40.61 & 30.43$\pm$8.82 & 10.00$\pm$6.85
\\

& \textbf{Ours}
& \textbf{108.85$\pm$15.46} & \textbf{16.16$\pm$2.59} & \textbf{28.46  $\pm$9.34}
\\

\bottomrule
\end{tabular}
 \end{adjustbox}
\end{table}
We calculate the average MPJPE, PA\_MPJPE, and PCK@15mm of the 14 upper body joints in {\it leave-one-subject-out experiments}. The results presented in TABLE  \ref{table:results_main} reveal that our approach remarkably surpasses the baseline model by average margins(take the average of the three actions) of 30\%, 45\%, and 184\% on MPJPE, PA\_MPJPE, PCK@15mm respectively. 

Furthermore, we evaluate the model's effectiveness on upper limb joints
pivotal to hand-object interactions. The the MPJPE and PA\_MPJPE of left and right wrist and elbow joints are summarized in Table~\ref{table:key_joints}. 
\begin{table}[!t]

\begin{center}

\caption{
Accuracy of upper limb key joint positions (in mm)
}
\label{table:key_joints}
\begin{adjustbox}{width=\linewidth}

\setlength{\tabcolsep}{0.5mm}{
\begin{tabular}{l | l |cc | cc }
\toprule
action
& 
method
& \multicolumn{2}{c}{MPJPE$\downarrow$} 
& \multicolumn{2}{c}{PA\_MPJPE$\downarrow$}
\\ 
& 
& \multicolumn{1}{c}{wrist} 
& \multicolumn{1}{c}{elbow}
%& \multicolumn{1}{c}{neck}
& \multicolumn{1}{c}{wrist} 
& \multicolumn{1}{c}{elbow}
%& \multicolumn{1}{c}{neck}
\\

\midrule

\multirow{2}*{driving}

& mmMesh  
&341.96$\pm$64.83&199.94$\pm$37.81%&138.525 
&103.35$\pm$25.63&66.50$\pm$18.48%&99.513 
\\

& \textbf{Ours}
& \textbf{119.46$\pm$25.71} & \textbf{114.30$\pm$14.16} %&106.140
& \textbf{38.95$\pm$6.98} & \textbf{28.45$\pm$6.20} %&70.469
\\
\midrule
\multirow{2}*{handreaching}

& mmMesh  
& 312.22$\pm$66.61 & 187.68$\pm$30.50 %&129.432 
& 96.44$\pm$21,47 & 64.44$\pm$16.39 %&91.447 
\\

& \textbf{Ours}
& \textbf{140.46$\pm$30.77} & \textbf{124.86$\pm$27.79} %&106.856
& \textbf{42.57$\pm$11.65} & \textbf{32.30$\pm$9.78} %&61.541 
\\
\midrule
\multirow{2}*{head}

& mmMesh   
& 377.14$\pm$102.65 & 226.84$\pm$56.17 %&167.079 
& 104.40$\pm$29.49 & 68.90$\pm$20.96 %&101.303
\\

& \textbf{Ours}
& \textbf{131.08$\pm$26.88} & \textbf{127.10$\pm$31.05} %&94.753 
& \textbf{38.70$\pm$9.41} & \textbf{27.96$\pm$6.84} %&67.681 
\\
\bottomrule
\end{tabular}
}

\end{adjustbox}
\end{center}
\end{table}

From the results, we can see that the average accuracy of wrist joints is lower than that of elbow and other upper body joints. This can be explained by the low RCS of hands, making them difficult to be captured by radars. However, SUPER considerably outperforms mmMesh in the estimation of both upper arm joints. Thus, we conclude that it is important to design a specific pipeline for SUB-HPE, and the inclusion of intensity features and the use of radar V are instrumental in improving the accuracy. 

% \input{tables/key_joints}
% \begin{figure}[th]
%   \centering
%   % Requires \usepackage{graphicx}
%   %\includegraphics[width=0.45\textwidth, height=0.2\textheight]{images/systemdiagram.jpg}
%   \includegraphics[width=\linewidth]{images/fig1.png}
%   \caption{\tobefilled{}.}
%   \label{fig:driving_key_joints}
% \end{figure}
%Furthermore, we evaluate the model's effectiveness on several pivotal joints deemed crucial for upper body pose estimation, especially in the context of in-vehicle driver monitoring. 
%Using the act of driving as an illustrative example, Figure~\ref{fig:dirving_key_joints} demonstrates that our model substantially outperforms mmMesh on both MPJPE and PA\_MPJPE and on all the selected key joints. 
%A comprehensive quantitative analysis was also conducted, the detailed results of which are presented in Table~\ref{table:key_joints}, showing our model's superiority not only on general joints but also on those challenging edge joints. 
\begin{figure}[th]
  \centering
  %\includegraphics[width=0.45\textwidth, height=0.2\textheight]{images/systemdiagram.jpg}
  \includegraphics[scale=0.42]{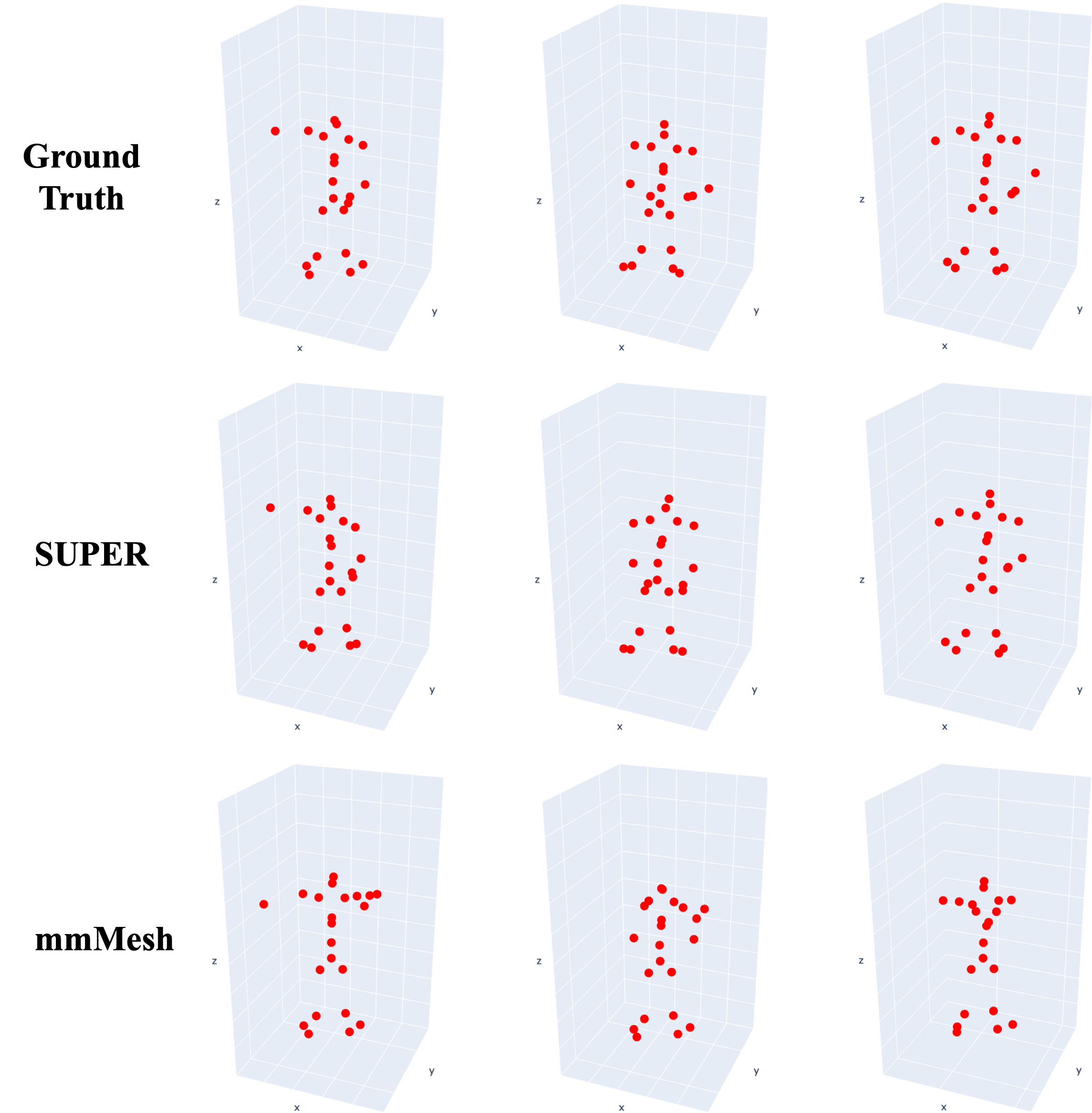}
  \caption{Constructed 3D poses in skeleton representation from SUPER, mmMesh and Ground Truth}
  \label{fig:pose_comparison}
\end{figure}

Examples of constructed 3D poses in skeleton representations using SUPER, mmMesh and Ground Truth can be found in Fig~\ref{fig:pose_comparison}.

\subsection{Ablation study}
\begin{table}[!h]
% \small

\begin{center}

\caption{
Results from Ablation Study
}
\label{table:ablation study}
\begin{adjustbox}{width=\linewidth}
\setlength{\tabcolsep}{1mm}{
\begin{threeparttable}
\begin{tabular}{l | l| c | c | c }
\toprule

action & information
& \multicolumn{1}{c}{MPJPE$\downarrow$} 
& \multicolumn{1}{c}{PA\_MPJPE$\downarrow$}
& \multicolumn{1}{c}{PCK@15mm$\uparrow$}
\\ 
\midrule

\multirow{3}*{driving}
& Doppler only   
& 237.33$\pm$26.72 & 44.78$\pm$2.29 & 0.37$\pm$1.46
\\
& intensity only
& 196.97$\pm$32.89 & 20.09$\pm$6.94 & 25.36$\pm$15.98 
\\
& Doppler+intensity
& 101.51$\pm$19.06 & 12.27$\pm$3.41 & 47.40$\pm$17.18
\\
\midrule

\multirow{3}*{handreaching}
& Doppler only   
& 266.28$\pm$36.64 & 48.49$\pm$3.74 & 0.08$\pm$0.91
\\
& intensity only
& 193.05$\pm$43.37 & 19.26$\pm$7.94 & 26.75$\pm$19.83 
\\
& Doppler+intensity
& 110.73$\pm$30.91 & 12.51$\pm$5.86 & 47.10$\pm$20.53
\\
\midrule

\multirow{3}*{head}
& Doppler only   
& 221.46$\pm$31.73 & 46.04$\pm$1.21 & 0.02$\pm$0.63
\\
& intensity only
& 190.71$\pm$44.41 & 19.05$\pm$7.47 & 27.68$\pm$18.90
\\
& Doppler+intensity
& 99.67$\pm$28.14 & 12.28$\pm$5.87 & 45.25$\pm$18.79
\\

\bottomrule

\end{tabular}

\begin{tablenotes}   
\scriptsize        
\item[*] The standard deviation in this table is calculated across the estimated position errors per joint and per frame.  
\end{tablenotes}

\end{threeparttable}
}
\end{adjustbox}

\end{center}

\end{table}
We further conduct ablation experiments to evaluate the effectiveness of Doppler and intensity point clouds in the training data. To do so, we only input the Doppler point cloud or the intensity point cloud and remove the respective branch in the backbone (Figure~\ref{fig:dnn_architecture}). Table~\ref{table:ablation study} reports the results from one test subject performing different actions. Clearly, neither intensity or Doppler point clouds alone is sufficient. Combining both sets of features leads to the highest accuracy. Somewhat interesting, between the two, intensity point clouds appear to be more informative. 
\section{Demonstrative Application}
\label{sect:app}
%In this section, we explore two downstream tasks to demonstrate the utility of SUPER: hand-object interaction and distraction detection via head pose estimation. Note that what is being presented acts as a proof-of-concept. Likely, more sophisticated methods can be implemented for the two tasks on top of SUB-HPE. 

In this section, we demonstrate the utility of SUPER through a downstream task that identifies hand-object interaction through SUB-HPE. Note that what is being presented acts as a proof-of-concept. Likely, more sophisticated methods can be implemented for the task on top of SUB-HPE. 

%
%\subsection{Hand-Object Interaction}
\begin{figure}[th]
  \centering
  %\includegraphics[width=0.45\textwidth, height=0.2\textheight]{images/systemdiagram.jpg}
  \includegraphics[scale=0.3]{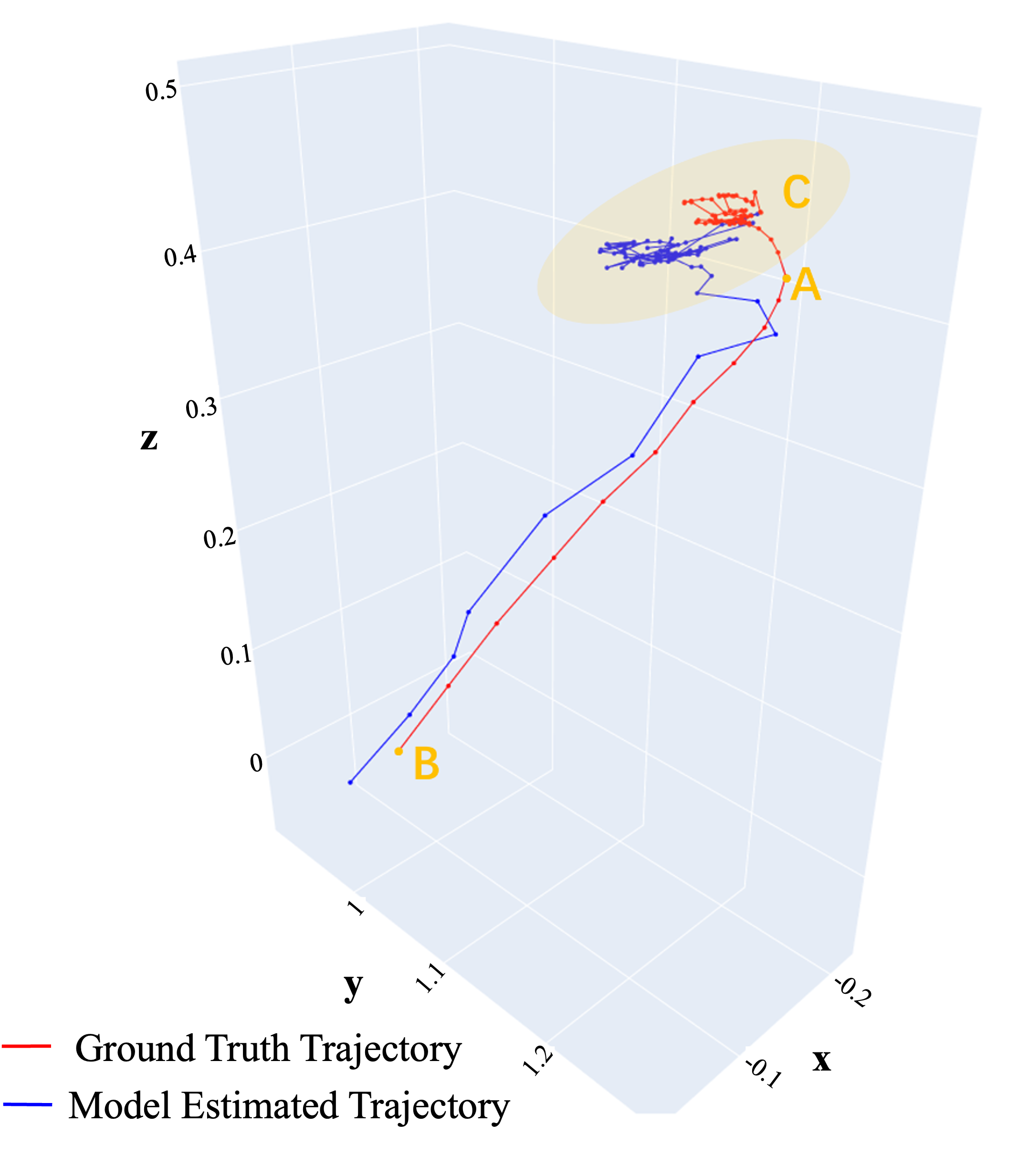}
  \caption{Visualization of the ground truth and model estimated wrist trajectories of a 4s sequence from a handreaching action. The units are in meters.}
  \label{fig:section6_1}
\end{figure}

In this task, the aim is to determine which objects in the 3D space one is interacting with by hands. Consider a motion sequence where a hand starts from some resting position, moves toward an object at known location, and then interacts with the object for a period of time. We transform the problem of object identification to a localization problem, namely, to determine whether one's hand (a wrist joint specifically) falls into the predefined bounding boxes around target locations for a sufficient amount of time. 

To test this idea, we first calculate the amount of displacement of a wrist joint during 1s windows in the ground truth trajectory. The intervals that the total displacement is less than a predefined threshold (100mm in the implementation) indicate either the initial rest position or the rendezvous point between the hand and a target object. We compute the centroid of the wrist joint positions in such intervals and test against the ground truth target locations. As an example, consider the ground truth and estimated trajectories as shown in Figure~\ref{fig:section6_1}. In this example, one's hand travels from $B$ to $A$ and then reaches a target location $C$. Although the estimated trajectory does not exactly coincide with the ground truth one, it can be observed as the hand approaches and stays around the target location, the estimated locations are close to $C$.

We conduct experiments on all subjects using the hand-reaching trials. The results show that in 88.80\% of the rest position or the rendezvous point intervals, the centroid of the estimated writ trajectory falls into a bounding box centered on the target location with a side length of 0.2m.

\section{Discussion and Conclusion}
\label{sect:conclusion}
In this work, we proposed SUPER, a pipeline for SUB-HPE. To address the challenges of nuanced upper body movements when seated, we obtained both intensity and Doppler point clouds by fusing data coherently from two radars with orthogonal orientations. Compared to a baseline method that only utilizes Doppler point clouds from a single radar, SUPER has superior performance in terms of all metrics for HPE. 

The current SUPER framework assumes the presence of a single subject and the knowledge of the ROI. It can be easily extended to multiple subjects and unknown ROIs when combined with a target detection component. The current model can also be trained with additional mesh errors in SMPL and a term reflecting temporal consistency and smoothness of human movements~\cite{zheng2023deep}. Doing so is expected to further improve the accuracy and realism of the inferred poses. 

Future research directions for mmWave-based SUB-HPE also include developing models that are robust to different deployment environments and the investigation of more downstream tasks. 

\bibliographystyle{IEEEtran}
\bibliography{dm}
\end{document}